\documentclass[letterpaper, 10 pt, conference]{ieeeconf}  % Comment this line out if you need a4paper

\IEEEoverridecommandlockouts                              % This command is only needed if 
\usepackage{hyperref}
\usepackage{graphicx}
\usepackage{amsmath}
\usepackage{cite}

\title{\LARGE \bf{OpenVR: Teleoperation for Manipulation}}

\author{Abraham George$^{1}$, Alison Bartsch$^{1}$, and Amir Barati Farimani$^{1}$
\thanks{$^{1}$With the Department of Mechanical Engineering,
        Carnegie Mellon University 
        {\tt\small \{aigeorge, abartsch, afariman\} @andrew.cmu.edu}}%
}

\begin{document}

\maketitle
\thispagestyle{empty}
\pagestyle{empty}

% Previous Header info: 

% \documentclass{article}

% % Language setting
% % Replace `english' with e.g. `spanish' to change the document language
% \usepackage[english]{babel}

% % Set page size and margins
% % Replace `letterpaper' with`a4paper' for UK/EU standard size
% \usepackage[letterpaper,top=2cm,bottom=2cm,left=3cm,right=3cm,marginparwidth=1.75cm]{geometry}

% % Useful packages
% \usepackage{amsmath}
% \usepackage{graphicx}
% \usepackage[colorlinks=true, allcolors=blue]{hyperref}

\begin{abstract}
Across the robotics field, quality demonstrations are an integral part of many control pipelines. However, collecting high-quality demonstration trajectories remains time-consuming and difficult, often resulting in the number of demonstrations being the performance bottleneck. To address this issue, we present a method of Virtual Reality (VR) Teleoperation that uses an Oculus VR headset to teleoperate a Franka Emika Panda robot. Although other VR teleoperation methods exist, our code is open source, designed for readily available consumer hardware, easy to modify, agnostic to experimental setup, and simple to use. 

\end{abstract}

\section{Introduction}

Teleoperation, or the remote control of a robot, is a widely used tool in robotics research. Although the ability for a human to take control of a robot is useful in a wide range of disciplines, this paper will focus on the application of teleoperation in machine learning (ML). One of the key challenges of robotic ML is the collection of expert trajectories to use for training Imitation Learning or Reinforcement Learning (with expert demonstrations). In recent years, many advancements have been made in the field of reinforcement learning, including work where the use of human demonstrations has proven vital \cite{Vargas2019Creativity, wang2023man, george2023minimizing}. Ideally, collecting these human demonstrations would be quick and easy to implement, and simple for the expert to use. This paper explores the use of an Oculus virtual reality headset, running an application developed in Unity, to teleoperate a Franka Emika Panda robot arm. 

One of the major challenges of using VR for teleoperation is that the view provided to the user's eyes must track with the user's head pose. A virtual camera that does not move in line with the headset, and do so at a fast enough rate (60 Hz), can induce VR sickness, a painful condition characterized by head-ache, nausea, and eye strain that can persist long after the VR session is completed \cite{Lee2017MOSKIT}. As such, any VR teleoperation setup must have a virtual camera that can quickly update with the head position of the user. There are two primary ways this can be accomplished if the robot's environment is observed using stationary cameras (as we do). Either the camera feeds can be projected onto virtual screens inside of the VR environment, appearing as images on screens in VR, or the features of the pertinent objects in the scene can be captured using computer vision, passed to the VR environment, then reconstructed as environment objects. In both these cases, the VR engine can move the virtual camera in tandem with the headset, adjusting the relative position of the screen or game objects, respectively. Because our teleoperation tool set is designed for reinforcement learning applications, we chose to take the object identification and reconstruction approach since we are interested in recording the relevant object features to assist with training the RL agent. 

Another challenge that guided our design process was creating a system that was not only easy to use, but also easy to set-up and modify. This consideration drove our decision to develop our software using Unity, and design it for the Oculus Quest 2. We chose to use Unity to develop the application because it is widely considered to be an easy-to-use development engine, and is the engine of choice among game developers, used by 61\% of developers \cite{irpan_gohil_tenboer_2021}. Similarly, the Oculus Quest 2 was chosen because it is the most popular XR (Virtual Reality + Augmented Reality) headset on the market, making up 66\% of all units shipped in Q2 of 2022 \cite{Chauhan2020XR}. Although the teleoperation program is built for the Franka Emika Panda robot arm, our implementation is designed such that any robot manipulator can be substituted in for the Panda Arm with minor alterations to the code.
Our code can be found at: \textit{\href{https://github.com/Abraham190137/TeleoperationUnity}{https://github.com/Abraham190137/TeleoperationUnity}}

\section{Related works}

Human demonstrations are necessary for many practical real-world applications including bootstrapping DRL, imitation learning and ML-based planning. There is a wide variety of techniques used to collect these human demonstrations that trade-off between ease of collection and demonstration quality. These techniques include visual demonstrations, leveraging key frames, corrective actions, kinesthetic teaching, and teleoperation, among others. Demonstrations collected purely from vision tend to be the most natural form of demonstration, as the demonstrator does not need to alter how they perform the task to provide the demonstration. Earlier works in this area, such as \cite{dillmann2010advances, welschedhold2016} tracked the trajectories of both the demonstrator and objects in the scene leveraging keypoint detectors. However, these methods were very brittle and struggled with the challenge of translating motions from human to robot embodiment, i.e. the correspondence problem. As the field of computer vision has advanced, significant improvements have been made in the area of visual demonstration. \cite{sermanet2018, sivakumar2022} each develop methods to learn robotic policies from unlabeled videos. \cite{ sivakumar2022} uses YouTube to collect a large amount of data to train a system that maps human hand motion to robot actions. However, despite these advancements, the data collected by visual demonstration remains less rich than other collection methods, as they provide estimated trajectories, but are unable to infer other helpful data modalities such as forces. Additionally, learning a mapping between human and robot embodiment currently is specific to a particular robot embodiment, i.e. their system can map from human hands to a human-like gripper, but would not be able to map to a parallel gripper.

Key frame demonstrations, where the teacher provides individual snapshots of the key points in the demonstration trajectory, maintain some ease of control while overcoming the correspondence problem. Key frame demonstrations have been used for simple humanoid robot tasks \cite{akgun2012trajectories}, imitating mouse gestures \cite{akgun2012keyframe}, and more recently, a more robust implementation to learn quadrotor control \cite{jin2022learning}. Keyframe demonstrations make it easier to control high DoF systems, as the user can control a subset of DoFs at a time. However, these demonstrations have a much lower frequency of information, and it is necessary to ensure that the critical key frames are included in the demonstrations. Additionally, keyframe demonstrations will not capture dexterous motions, limiting the applicability to certain tasks. 

Corrective actions are another form of demonstration that attempt to balance the tradeoff between ease of collection and quality of data, where the robot already has a base policy, and the demonstrator provides instruction by physically correcting the robot's trajectory. This is a very intuitive form of demonstration, as this is very similar to how one may teach a child a new skill. However, this requires there to be a pre-defined base skill for the robot to execute, and will likely need multiple runs and multiple corrective actions to get a full, rich demonstration \cite{bajcsy2018learning, gutierrez2019learning}. In recent work, \cite{mehta2022unified} presents a pipeline for autonomous assembly that combines kinesthetic demonstrations with corrective action and preference selection to learn a reward function.

Kinesthetic teaching is a classic demonstration technique where the teacher physically moves the robot to perform the task. Kinesthetic teaching has been used to learn motor primitives \cite{su2016learning}, training dynamic movement primitives \cite{ragaglia2018robot,gavspar2018skill}, techniques that handle multi-modal demonstration inputs \cite{le2021forceful, shao2021concept2robot, stepputtis2022system}, as well as in numerous imitation learning and deep reinforcement pipelines \cite{johns2021, gams2022manipulation, kawamura2020forceadjustement}. Kinesthetic teaching provides very high quality data, as there will be ground truth state and proprioceptive information for each step of the demonstration trajectory. However, this is a very cumbersome and unintuitive demonstration technique; the teacher often needs to fully re-learn how to perform the task. This makes collecting demonstrations with kinesthetic teaching very expensive. Additionally, due to the nature of this technique where the teacher physically moves the robot, we typically cannot collect information from any other sensors, e.g. force, temperature, or sound, which can be incredibly valuable for certain applications.

Teleoperation provides the highest quality demonstration data, as this method directly provides the action for each state along the trajectory.  In addition, teleoperation can also accurately collect data from other sensor inputs, such as force/torque measurements, sound, or temperature. However, classical teleoperation techniques still require the demonstrator to significantly change how they would perform the task, making it very time-consuming to collect demonstrations. Due to this challenge, there is a wide variety of teleoperation methods that attempt to make demonstration collection easier and more intuitive. For example, \cite{yang2016neural} uses a joystick, \cite{jin2016multi} uses an infrared sensor, \cite{fang2017robotic, edmonds2017} use a wearable device, \cite{meklander2021} uses phone-based teleoperation, \cite{fang2015} uses glove-based teleoperation, and \cite{mandlekar2018roboturk} created a phone-based crowd-sourcing platform for teleoperation. Each of these teleoperation methods improves usability for the demonstrator but still requires some re-learning of how to perform the task given the control scheme. 

Virtual reality (VR) teleoperation is able to overcome the classical trade-off between ease of collection and demonstration quality by making teleoperation much more intuitive to use. However, existing VR teleoperation methods are developed for specific robotic systems, such as drones \cite{walker2019}, or 5-fingered hand \cite{arunachalam2022}, have not been made publicly available for other research groups to use \cite{zhang2018}, or do not offer the capabilities of performing the tasks fully in simulation without real-world image input \cite{whitney2018}. Although these alternative VR teleoperation systems already exist, in this work we present a teleoperation pipeline that is open source, easy to modify, agnostic to experimental setup and robot type, and simple to use.

\section{Methodology}
\subsection{Overall Pipeline}

\begin{figure}[thpb]
  \centering
  \includegraphics[width=\linewidth]{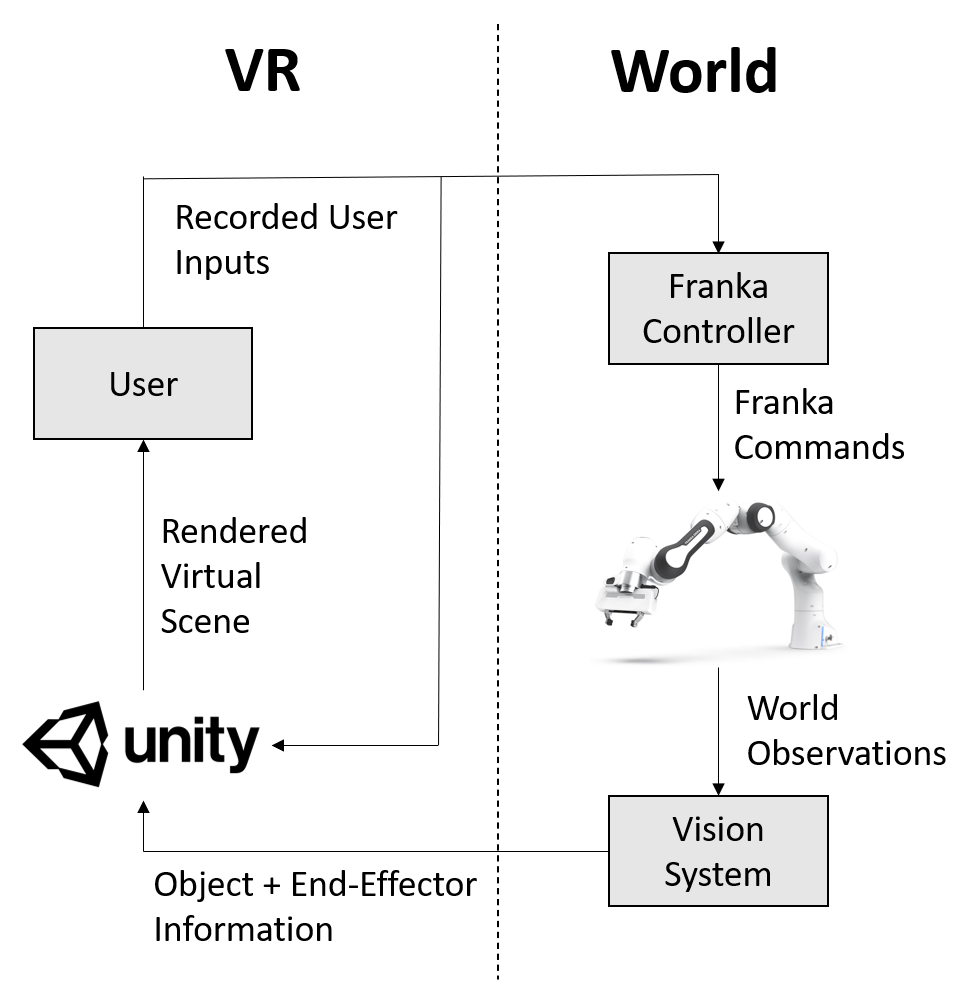}
  \caption{\label{fig:pipeline}
  Overall pipeline of the teleoperation system. The Unity application (running on an Oculus headset) renders a virtual scene that is presented to the user. The user's inputs (in the form of hand-held controller movements) are recorded by Unity and sent to the Franka-Emika robot controller. The controller executes these commands, and the vision system observes the resulting change in the environment. The observed environmental information, along with the state of the end-effector, is sent to Unity, which updates the rendered virtual scene.}
\end{figure}

The general outline of the teleoperation system is as follows: in virtual reality, the user sees a rendered version of the environment the robot needs to interact with, along with a sprite representing the robot's end-effector which tracks the location of the real robot's hand. The user controls the robot by moving the Oculus VR headset's hand-held controller. The location of the controller, along with the goal width of the parallel plate gripper (adjusted with the controller's joystick) is sent across a UDP socket to the robot's base station, which runs the robot controller script. Here, the goal location is passed into a trajectory following script, causing the robot to move toward the end-effector's goal pose. Next, RealSense cameras capture a video feed of the scene (where the robot has acted). This video data is then parsed to get object data, which is passed across a UDP socket to the application running on the Oculus headset along with the updated pose of the robot hand. Here, the object data is used to update the states of the current objects in the scene, remove objects that are no longer needed, and generate new objects. This process repeats for the duration of the teleoperation, showing the user a rendering of the environment, using the user's input to control the robot, and updating the rendered environment with the observations of the real world.

\subsection{Unity Environment}
\begin{figure}[thpb]
  \centering
  \includegraphics[width=\linewidth]{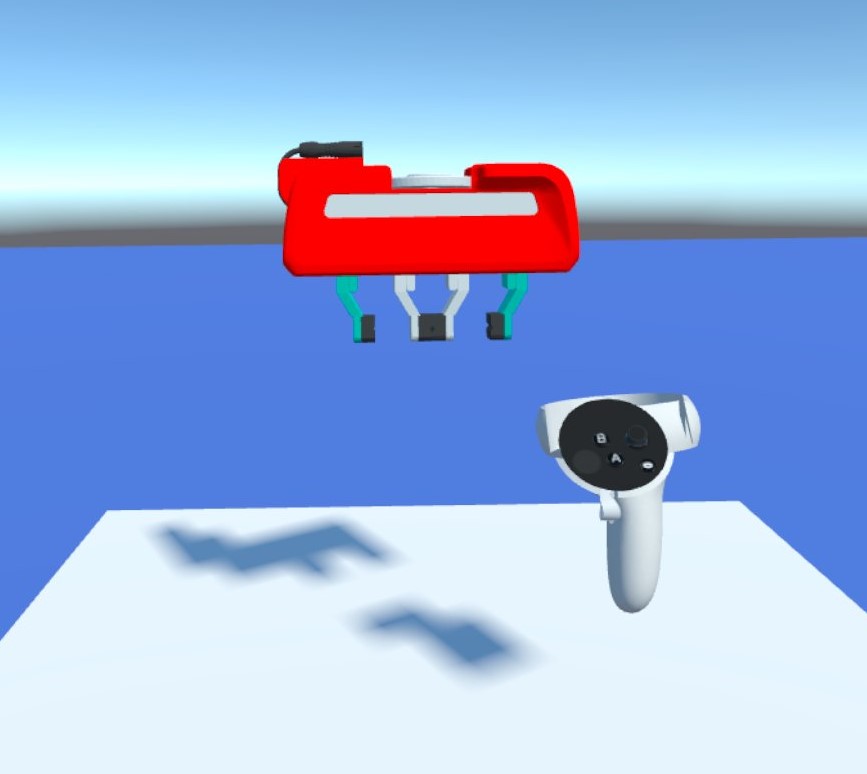}
  \caption{\label{fig:controller_feedback} Unity end-effector rendering (paused). The sprite tracks the location of the Franka-Emika Panda hand. An extra set of green grippers is used to show the 'goal' location of the gripper. The user directly controls the position of these green fingers, which is then passed to the FrankaPy controller. The rendering of the Oculus controller gives the user position feedback. Since the simulation is paused, the gripper is highlighted in red.}
\end{figure}

The virtual reality environment used for teleoperation was developed in Unity, using the Oculus plug-in and Unity’s robotic package’s URDF importer. The Oculus plug-in allows for easy development of VR applications, with a built-in head-tracking virtual camera, controller mapping, and additional development features. The URDF importer was used to recreate the Franka-Emika Panda end effector in the Unity environment. The URDF code was sourced from \cite{UnityRobticsHub}. The imported end-effector model is used as a non-kinematic sprite, providing visual feedback to the user about where the end-effector is without physically interacting with the rest of the Unity simulation. Therefore, the URDF imported model can be replaced with any accurate model of the chosen robot’s end-effector. An image of the Unity environment can be seen in Fig. \ref{fig:controller_feedback}

The objects in the Unity environment can be broken into two categories: permanent objects and observed objects. Permanent objects make up the constant elements of the environment the robot is working in, such as the floor, the table the robot is mounted to, and the base of the robot itself. This category also includes all of the permanent Unity assets needed to properly render in and interact with the world, such as lighting, the VR camera, and hand trackers. The rendered copy of the robot’s end effector is also included in this category. The observed objects consist of all environmental elements which are identified, classified, and tracked by the vision system. In the unity environment, all of the permanent objects are included in the scene from the unity editor and can be modified there. The observed objects, however, are script-generated. When the vision system identifies a new object, it sends the object’s details to the Unity application, where a script generates the desired object and places it in the scene. The generated objects can then be moved, altered, or destroyed by subsequent messages from the vision system.

The user provides input for the control of the end-effector through the location sensors on the oculus hand-held controllers and the controller's joystick. The end-effector tracks the user's right hand as it moves through space in the virtual environment, and the width of the parallel plate gripper is controlled by the right-hand joystick (up opens the gripper, down closes it). Due to the low frequency of the gripper control, the user does not directly control the gripper. Instead, the joystick adjusts the position of a 'goal' gripper, and the goal position is sent to the Franka controller. In addition to mitigating latency concerns, this setup also allows the user to close the goal gripper on objects further than physically possible, ensuring the end-effector maintains a strong grip.

To prevent the user from sending a command to the robot that would move the end-effector outside of its designated workspace, virtual walls were added to the Unity simulation. If the goal pose of the end-effector causes any part of the end-effector to violate these walls, the pose command is prevented from being sent to the Franka controller. Whether a desired pose will cause the end-effector to leave the workspace is determined with a convex polygonal bounding box. To check this, each vertex of the bounding polygon is transformed using the desired pose's rotation and transition, and then each point is compared to the virtual walls. If an end-effector other than the franka-emika panda gripper is used, then the vertex points can be modified accordingly.

The unity interface also has a pause feature, which is toggled by the A button on the Oculus controller. When control is paused, Unity stops updating the goal end-effector location and gripper width, allowing the user to move their hands freely. When paused, the end-effector rendering in Unity changes color to red to alert the player. A major issue with freezing the end-effector updates (either due to the user pausing the game or violation of the virtual walls) is that the rendered gripper (which tracks the actual position of the end-effector) does not provide any visuals as to where the hand is being commanded to go (the location of the Oculus controller). To give the user feedback about where their hand is, a controller sprite was added to the Unity world, which directly tracks the location of the right oculus controller. This sprite is semi-transparent, with its opacity proportional to the distance between the user's hand position and the robot's end-effector location. This relationship causes the controller sprite to be invisible during normal operation when the end-effector is successfully tracking the user's inputs, but become fully visible when this discrepancy becomes large, such as when the user is violating the virtual walls. The gripper is also set to fully opaque when the game is paused.

In order for the communication system to work, the FrankaPy controller must know the Oculus's IP address, and Unity needs to know the IP address of the computer running the FrankaPy controller. To aid in the entering and reading of this information, a start screen was created to display the Oculus's IP address and record the IP address of the FrankaPy controller. This screen appears when the application is first booted and can be reached anytime during operation by pressing the B button.

\subsection{FrankaPy Robot Control}

The control module for the Franka robot arm is built on FrankaPy, a python wrapper for controlling the Franka Emika Panda arm \cite{zhang2020modular}. From this library, we use four primary commands: \textit{go-to-pose}, \textit{go-to-gripper}, \textit{get-pose}, and \textit{get-gripper-width}. These commands move the end-effector to the desired pose, set the parallel plate gripper width, return the end-effector's location, and return the gripper's current width, respectively. Although we use FrankaPy, any other robotic control method can be substituted, so long as it has similar commands. At each update step, the FrankaPy controller receives a goal gripper location and width from Unity. These commands are passed into the respective go-to commands, and the current pose and gripper width are retrieved using the respective get commands and passed to Unity.

\begin{figure}[thpb]
  \centering
  \includegraphics[width=\linewidth]{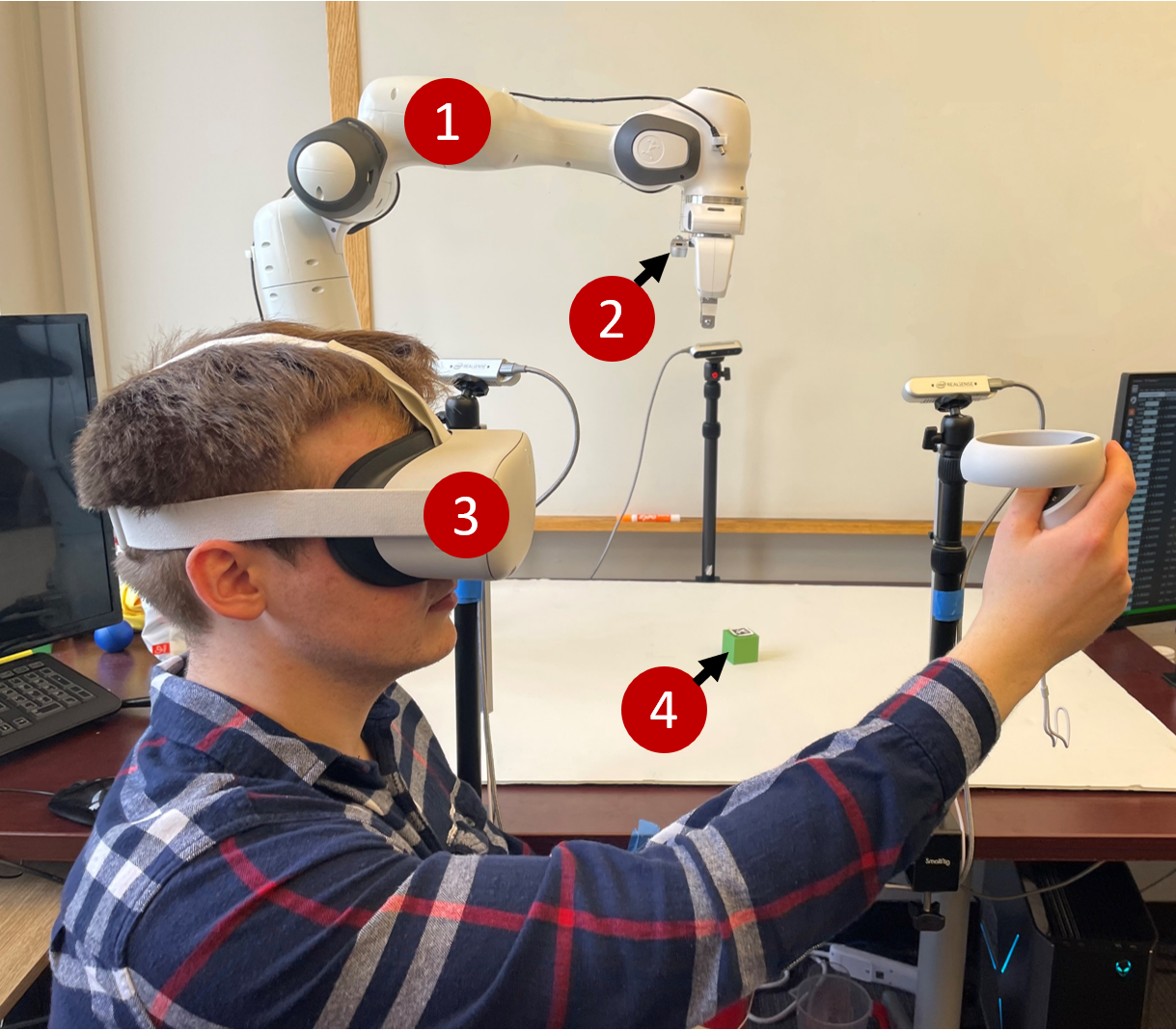}
  \caption{\label{fig:hardware_setup} View of the hardware setup. 1) Franka Emika Panda robot, 2) Intel RealSense D415 camera, 3) Oculus headset, 4) object in the environment.}
\end{figure}

The vision system works in tandem with the controller, collecting information about the objects in the robot's environment. Our system uses a wrist-mounted camera and a static camera that views the entire table (both RealSense D415), although any set-up that can identify and determine the location of objects in the scene will work. Likewise, we use April tags for object detection and localization, but this approach is not required. An image of our setup can be seen in Fig \ref{fig:hardware_setup}. In our setup, each object we wish to track has a corresponding April tag, which the vision system can use to reference the object's features, such as shape, size, color, and name. Additionally, the cameras determine the April tag's location and rotation in the world frame. From these measurements, the system calculates the pose of the object. This pose, along with the object velocity (calculated using previous pose measurements), is then passed to Unity. If the object is detected entering the scene, then an object creation message containing the object's features is passed to Unity, and if an object leaves the scene, then a command to delete the object is sent to Unity.

\subsection{Communication Method}

\begin{figure}[thpb]
  \centering
  \includegraphics[width=\linewidth]{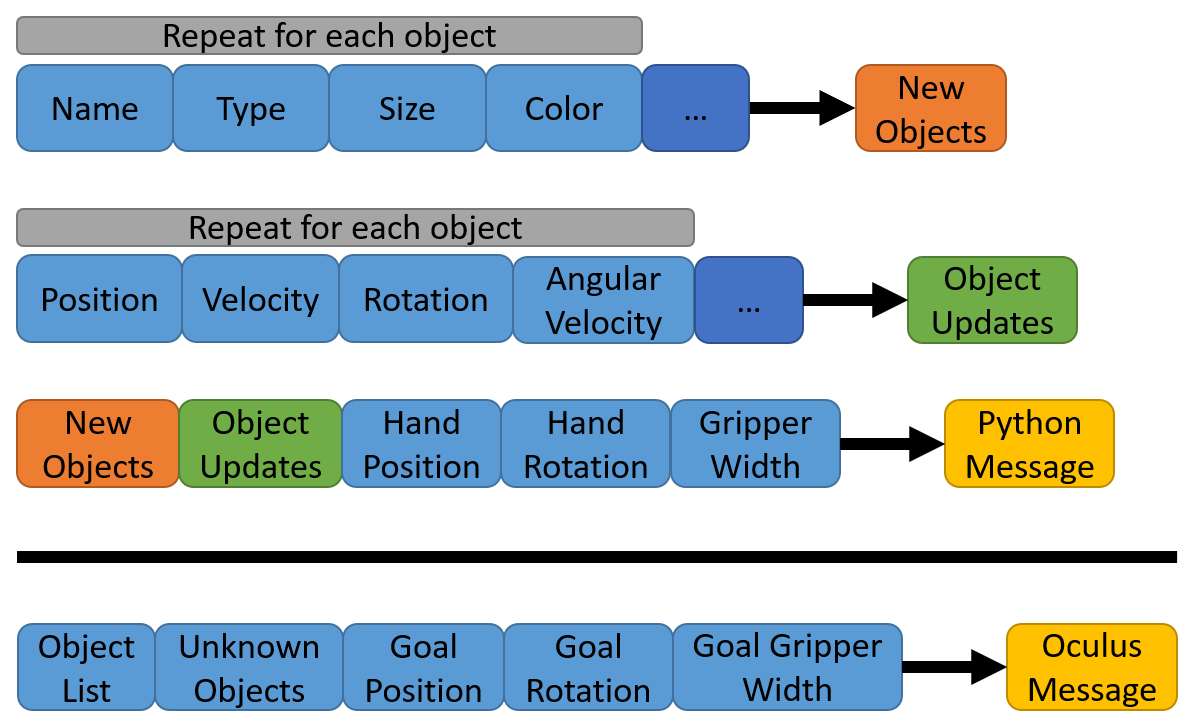}
  \caption{\label{fig:message_graphic} Diagram showing the components of the communication messages from the python based controller to the Unity based VR application (top), and from the Unity application to the python controller (bottom)}
\end{figure}

To communicate between FrankaPy and Unity, we used the UDP Socket pipeline developed by \cite{Youssef2022Communication}. Each Unity update (frame), the UDP socket packages a message and sends it to the Python controller, where the message is parsed and used for control of the robotic arm. Similarly, at each of our controller code's updates, the controller sends a message across the UDP socket to Unity. Here, the message is parsed and used to update the Unity scene. A diagram of these messages can be seen in Fig \ref{fig:message_graphic}

The message that the unity application receives contains information on the pose of the gripper, the updated pose and velocity (both transitional and rotational) of the objects in the scene, and occasionally contains create object and delete object commands. The hand pose information is used to update the end-effector sprite, directly setting the pose of the hand and fingers. Similarly, the object pose and velocity message is used to directly set the pose and velocity of the objects in the scene. To keep track of objects in both our controller and unity, each object is assigned a unique key upon creation, which is passed along with the kinematic information in the update message. In unity, these keys are used to reference the corresponding object's rigid body from the object dictionary. If Unity receives an update message for an object that doesn't exist (if the received object key is not in the object dictionary), then Unity will skip that object update and add it to the list of unknown objects, which is then sent back to the controller script.

In addition to updating existing objects, new unity objects need to be created when objects enter the robot's workspace and old unity objects need to be deleted when the physical items they correspond to leave the robot's work space. To accomplish these tasks, our Python controller sends create object and destroy object messages. The create object message includes the type of object to be created (for example, 'block'), the object's key, and all of the information needed to generate that item. This information is then passed to the object-generating script for that type of object. For example, if the object type is 'block', then the object's parameters are passed into the block generation script. This script generates an appropriate kinematic rigid body without a collider and with gravity disabled. These settings prevent the objects from interacting with any other objects in the environment. Finally, the new object's rigid body, along with its key, is saved to the object dictionary, where it can later be accessed to update the object's pose and velocity. In addition to creating objects, the communication protocol can delete old objects. This is done with the delete object command, which deletes the specified object from the unity scene and removes it from the object dictionary. If the Unity script receives a command to generate a new object with a key that is already present in the object dictionary, it will first delete the existing object with that key, then execute the create object command.

After processing the update message, Unity sends a response update message to the controller script. This update message includes the goal end-effector position and the list of known object keys. The goal end-effector positions are passed into the go-to pose and go-to gripper commands, and the Unity key list is compared with a list of all of objects the cameras detected. If an object is present in both lists, then the Python script sends an object update message. If the cameras detect an object that is not in the Unity list, a create object message is sent instead. If the cameras detect that an object in the Unity list has left the scene, then a delete object command is sent.

\subsection{Virtual Environment Teleoperation}
\begin{figure}[thpb]
  \centering
  \includegraphics[width=\linewidth]{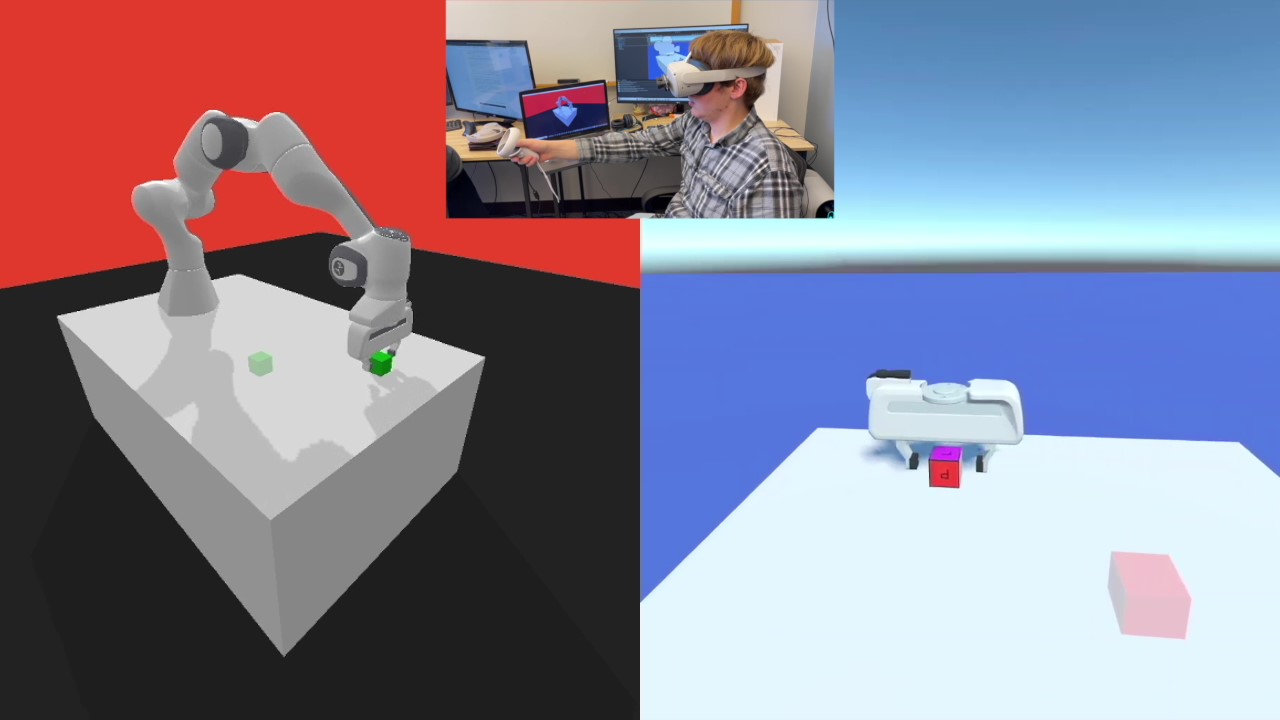}
  \caption{\label{fig:simulation_demo} View of the the Panda-Gym simulation teleoperation setup. The image on the left shows a rendering of the Panda-Gym simulation, the right shows the corresponding view in the VR headset, and the top image shows the user controlling the VR system. Here, the user's motions (top) are captured in the VR environment (left), and the corresponding control messages are sent to the simulation (right).}
\end{figure}

In addition to teleoperating the Franka Emika Panda robot, we developed a simulation-based alternative, which controls a virtual Franak Emika robot in Panda Gym, a pybullet-based simulator \cite{gallouedec2021}. We developed this alternative to allow individuals without robotic hardware to use our system to do simulation-based work, to provide us with a method to analyze our software's performance while isolating the limitations of the teleoperation software from those of the hardware, and to demonstrate the versatility of our methodology by implementing the teleoperation system with a different control interface. The simulation-based teleoperation system works similarly to the hardware implementation, except the robotic control and vision tasks are done in simulation, not on hardware. The robot control script is nearly identical to the control script from the hardware implementation, except the FrankaPy \textit{go-to-pose}, \textit{go-to-gripper}, \textit{get-pose}, and \textit{get-gripper-width} commands are replaced with equivalent Panda-Gym commands. The object detection process was more significantly changed since the ground truth positions of the objects are known in the simulator. An image of the simulator scene can be seen in Fig. \ref{fig:simulation_demo}.

\section{Additional Gripper: Robotic Hand}
\begin{figure}[thpb]
  \centering
  \includegraphics[width=\linewidth]{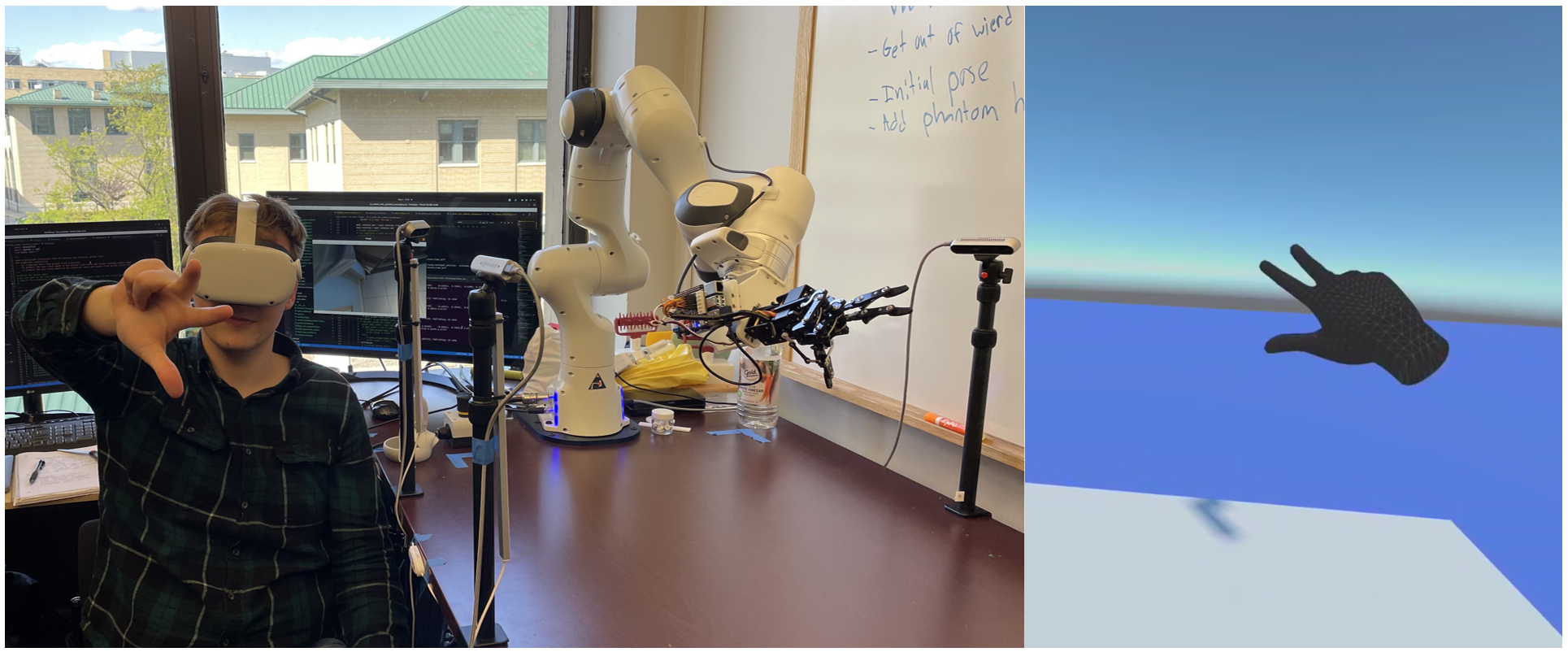}
  \caption{\label{fig:hand_teleop} Five fingered hand teleoperation setup. On the right is a view of the user and the Franka Emika Panda robot with the mounted hand, and on the left is the user's view in VR.}
\end{figure}

To help illustrate the process of altering the gripper and to increase the functionality of our teleoperation setup, we created a modified version of our original design, replacing the Franka Gripper with Hiwonder's uHandPi Robotic Hand. This hand, which seeks to mimic the motions of a human hand, has six degrees of freedom, one for each finger plus the wrist, and is controlled by a Raspberry Pi. An adapter to mount the hand onto the arm was 3D printed, and the hand was installed onto the Franka arm, as seen in Fig. \ref{fig:hand_teleop}. To control the hand, we used the Oculus headset's built-in hand tracking to track the user's fingers. The amount each finger bends is sent across our UDP socket to the Franka controller, where it is passed to the Raspberry Pi, which moves the fingers accordingly. In order to determine where the hand should move, we once again used the built-in Oculus hand tracking. Luckily, the command in Unity for returning the position of the controller returns the position of the user's hand when hand-tracking is enabled, so our existing structure could be used, with the small addition of a pose offset to align the robotic hand with the user's hand. Once these alterations were made, we were able to teleoperate the hand with finger level control (Fig. \ref{fig:hand_teleop})

\section{Results}
\subsection{Block Stacking Demonstration}
\begin{figure}[thpb]
  \centering
  \includegraphics[width=\linewidth]{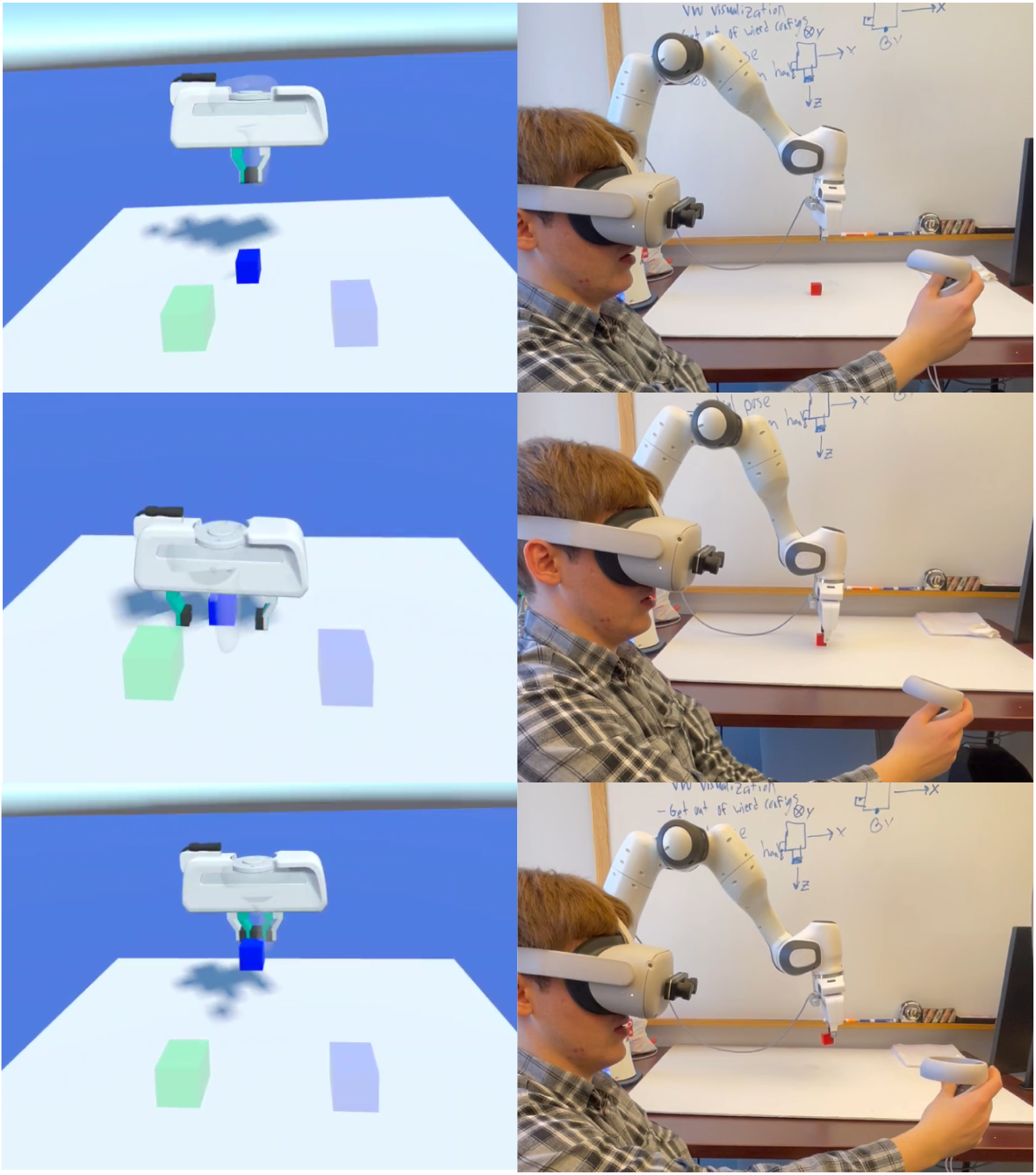}
  \caption{\label{fig:block_pick} Pick-and-place demo. The user opens the gripper, moves the gripper to the block, closes the gripper, and then carries the block to the goal location. By shutting the goal grippers (green) all the way, the user can command the gripper to constantly apply a force to the block, keeping it from falling.}
\end{figure}

To verify the functionality of our setup and illustrate the key aspects of the teleoperation setup, we performed a block pick and place task. In this task, the user moves the end-effector to a block, opens the gripper, picks up the block, and moves it. Fig. \ref{fig:block_pick} shows this test. Here, the gripper follows the user's motions throughout the grasping action. To ensure the block is held securely, the user sets the goal gripper width, shown by the green fingers, to fully closed, causing the controller to constantly apply force to the fingers. Also, since the user controls the goal gripper, not the actual gripper, they can achieve this constant application of force without constantly providing input. The user could then release the block by opening the goal gripper, causing the actual gripper to let go of the block.

\subsection{Latency}
To examine the latency of our system, we ran two tests to determine the operating frequency of our system. The first one was run using the simulated environment, where latency is dominated by the connection between the controller and the Oculus, and the second was run with hardware, where latency is dominated by the Franka controller. In both of these tests, we had four objects in the scene. From these tests, we found the teleoperation system operated at an average of 71.3 Hz in the simulated environment, with a 1\% low of 47.6 Hz. On hardware, the teleoperation system operated at 36.2Hz, with a 1\% low of 6.4 HZ. The significant decrease in the operating frequency when compared to the simulation suggests that our limitations are mainly imposed by the communication delay between the control script and the low-level Franka controller. However, an average of 36.2 Hz is more than sufficient, since the script only updates goal positions. The low-level control of the robot is performed by the FrankPy library at 1kHz. That being said, if the user needed a higher refresh rate, they could achieve significant improvement without altering the teleoperation setup by simply changing the robot control scheme. 

\subsection{Communication Overload}
\begin{figure}[thpb]
  \centering
  \includegraphics[width=\linewidth]{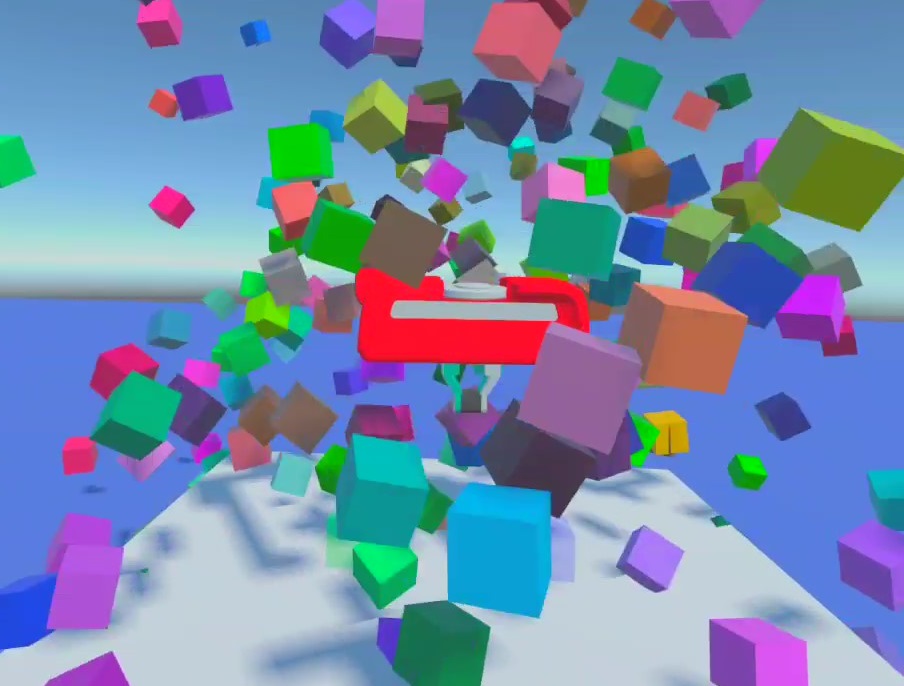}
  \caption{\label{fig:cubes!} Image of the communication overload test. In this test, the maximum number of objects the communication system could handle was examined by adding randomly moving cubes to the environment until the communication system threw an error.}
\end{figure}

Because our teleoperation system bundles all of the update information for the scene in one message, the size of these messages and the resources taken to process the message can become a limiting factor. To examine this concern, we ran an object overload test. For this test, we generated a new block object four times a second, with a random color, location, velocity, and angular velocity, and ran the simulation until an error occurred. During the simulation, the blocks moved around randomly in the workspace (Fig. \ref{fig:cubes!}). We found that our system could handle on average 240 objects without any issues. Across our 20 tests, the lowest number of object instances that caused the system to have an error was 219. These numbers suggest that as long as less than 200 dynamic objects are being communicated, our system is sufficiently robust. 

\section{Conclusion}
The teleoperation system detailed in this paper allows for Virtual Reality teleoperation of a Franka Emika Panda robot using inexpensive readily available hardware and open-source tools. The system is designed to be easily modified to fit a wide variety of use cases. To illustrate this aspect and to provide additional functionality, our code includes capabilities for use with a simulated Franka Emika Panda, and for a five-fingered hand, operated using hand tracking. Additionally, the inclusion of native support for an open-source simulator (Panda-Gym) allows anyone with an Oculus headset to perform robotics research using teleoperation.

\addtolength{\textheight}{-18cm}   % This command serves to balance the column lengths

\bibliographystyle{IEEEtran}
\bibliography{sample}

\end{document}